\documentclass[journal]{IEEEtran}
\ifCLASSINFOpdf
\else
\fi

\usepackage{graphicx}
\usepackage{comment}
\usepackage{amsmath,amssymb} 
\usepackage{color}

\graphicspath{{figures/}}
\usepackage[breaklinks=true,bookmarks=false]{hyperref}
\usepackage{epsfig}
\usepackage{graphicx}
\usepackage{amsmath}
\usepackage{amssymb}
\usepackage{authblk}
\usepackage{makecell}
\usepackage{algpseudocode,algorithm}
\usepackage{verbatim}
\usepackage{color}
\usepackage{booktabs}
\usepackage{siunitx}

\begin{document}
%
\title{EnAET: A Self-Trained framework for Semi-Supervised and Supervised Learning with Ensemble Transformations}
%
%
%

\author{Xiao Wang, Daisuke Kihara, Jiebo Luo~\IEEEmembership{Fellow,~IEEE,}
         Guo-Jun Qi,~\IEEEmembership{Senior Member,~IEEE}
\thanks{The work was done while X.~Wang was interning at Futurewei Technologies. X.~Wang was with Department of Computer Science, Purdue University, West Lafayette, 47906, USA.}
\thanks{D.~Kihara was with Department of Computer Science and Department of Biological Sciences, Purdue University, West Lafayette, 47906, USA. }
\thanks{J.~Luo was with University of Rochester, Rochester, 14627, USA.}
\thanks{G-J.~Qi was with was with the Futurewei Seattle Cloud Lab, Seattle, WA, 98006, USA. Email: guojunq@gmail.com.}
\thanks{Manuscript received 1 June, 2020; revised 20 Oct, 2020; accepted xx. (Corresponding author: Guo-Jun Qi)}
}
%
%

\markboth{IEEE Transactions on Image Processing,~Vol.~14, No.~8, August~2020}%
{Wang \MakeLowercase{\textit{et al.}}: EnAET: A Self-Trained framework for Semi-Supervised and Supervised Learning with Ensemble Transformations}
%



\maketitle

\begin{abstract}
Deep neural networks have been successfully applied to many real-world applications. However, such successes rely heavily on large amounts of labeled data that is expensive to obtain. Recently, many methods for semi-supervised learning have been proposed and achieved excellent performance. In this study, we propose a new EnAET framework to further improve existing semi-supervised methods with self-supervised information. To our best knowledge, all current semi-supervised methods improve performance with prediction consistency and confidence ideas. We are the first to explore the role of {\bf self-supervised} representations in {\bf semi-supervised} learning under a rich family of transformations. Consequently, our framework can integrate the self-supervised information as a regularization term to further improve {\it all} current semi-supervised methods. In the experiments, we use MixMatch, which is the current state-of-the-art method on semi-supervised learning, as a baseline to test the proposed EnAET framework. Across different datasets, we adopt the same hyper-parameters, which greatly improves the generalization ability of the EnAET framework. Experiment results on different datasets demonstrate that the proposed EnAET framework greatly improves the performance of current semi-supervised algorithms. Moreover, this framework can also improve {\bf supervised learning} by a large margin, including the extremely challenging scenarios with only 10 images per class. The code and experiment records are available in \url{https://github.com/maple-research-lab/EnAET}.
\end{abstract}


%
\IEEEpeerreviewmaketitle

\section{Introduction}
%
%
%
%
\label{sec:intro}

Deep neural network has shown its sweeping successes in learning from large-scale labeled datasets like ImageNet~\cite{deng2009imagenet}. However, such successes hinge on the availability of a large amount of labeled examples~\cite{qi2016hierarchically,hua2008online,qi2008correlative} that are expensive to collect. Moreover, deep neural networks usually have a large number of parameters that are prone to over-fitting. Thus, we hope that semi-supervised learning can not only deal with the limited labels but also alleviate the over-fitting problem by exploring unlabeled data. In this paper, we successfully prove that both goals can be achieved by training a semi-supervised model built upon self-supervised representations.
\par Semi-Supervised Learning (SSL)~\cite{chapelle2009semi,grandvalet2005semi,gong2016multi,wu2017semi,nie2017auto,qi2018global,wang2012recommending,zhao2018adversarial} has  been extensively studied due to its great potential for addressing the challenge with limited labels. Most  state-of-the-art approaches can be divided into two categories. One is confident predictions~\cite{grandvalet2005semi,miyato2018virtual,lee2013pseudo}, which improves a model's confidence by encouraging low entropy prediction on unlabeled data. The other category imposes consistency regularization~\cite{cirecsan2010deep,laine2016temporal,sajjadi2016regularization,tarvainen2017mean} by minimizing discrepancy among the predictions by different models. The two approaches employ reasonable objectives since good models should make confident predictions that are consistent with each other. Also, apart from image processing, semi-supervised learning ideas have achieved great success in other areas~\cite{zhuang2017label,dupre2019improving,tao2017scalable,turian2010word,papandreou2015weakly,cheplygina2019not}.
\par 

In self-supervised learning, researchers have explored several methods~\cite{zhang2019aet,he2020momentum} to fully utilize the latent information in images. In our previous work~\cite{zhang2019aet,Qi_2019_ICCV}, we utilized the image transformations as the supervising signal to train encoders, which have achieved promising performance. Instead of only exploring the information of labels to improve semi-supervised learning performance, we believe that the latent information carried by images can also contribute to the semi-supervised learning. As addressed in the  literature~\cite{hinton2011transforming,gidaris2018unsupervised,zhang2019aet}, a representative encoder should also recognize an object even if it is transformed in different ways. With deep networks, this is usually achieved by training the model with augmented labeled data. However, that will make the trained feature influenced by the label information and thus prevent the features from being more representative. Instead, inspired by those self-supervised methods, we used the transformations as the signal to supervise those transformed images to overcome the information loss introduced by the labels during semi-supervised training.

To this end, we will present an Ensemble of Auto-Encoding Transformations (EnAET) framework to self-train semi-supervised classifiers with various transformations by integrating self-supervised representations~\cite{zhang2019aet} to augment semi-supervised learning performance.  Our contributions are summarized as follows:
\begin{itemize}
    \item We propose the first method that employs an ensemble of both spatial and non-spatial transformations  to train a semi-supervised network in a self-supervised fashion over both labeled and unlabeled data.

    \item We apply an ensemble of AutoEncoding Transformations to learn robust features under various transformations, and improve the prediction consistency on the transformed images by minimizing their KL divergence. The proposed framework is a general framework that can be readily integrated with and improve any semi-supervised learning method.

    \item We demonstrate that the EnAET can greatly improve both the semi-supervised and supervised baselines by large margins through  integrating the self-supervised representations. Moreover, we find out the number of labeling data can be further reduced to achieve desired performances. 

    \item We use the same hyper-parameters across different datasets without over-tuning them in the experiments, further demonstrating the model's excellent generalization ability and practical value.
\end{itemize}
\par The remainder of the paper is organized as follows. We briefly review the related work in semi-supervised learning and self-supervised learning in Section \ref{sec:relatedwork}. We present our algorithm EnAET in Section \ref{sec:alleaet}.  To prove our framework's effectiveness and stability, extensive experiments related to supervised learning and semi-supervised learning are described in Section \ref{sec:experiments}. Finally, we conclude in Section \ref{sec:conclusion}.

\section{Related Work}
\label{sec:relatedwork}
In this section, we review both semi-supervised and self-supervised learning approaches in the literature.
\subsection{Semi-Supervised Learning (SSL)}
\par  Semi-Supervised learning~(SSL) is an approach in machine learning that performs training under the scenario where a large amount of unlabelled data and a small amount of labelled data are available.  Considering the fact that annotated data is expensive to collect and unlabelled data is easy to collect, it is necessary to develop methods that can learn from a small amount of annotated data.
\par A wide variety of SSL methods have been developed in literature, which aims to learn from limited number of annotated data. For example, Teach-Student Models~\cite{rasmus2015semi,laine2016temporal,tarvainen2017mean,miyato2018virtual} constitute a large category of SSL, which is closely related to the proposed model and is proposed based on the assumption that two online models can learn from each other through their predicted probabilities. Inspired by unsupervised learning~\cite{qi2019small,qi2016joint} performance on feature representation,  supervision information is incorporated into the variational auto-encoders to learn various semi-supervised classifiers ~\cite{kingma2014semi,narayanaswamy2017learning,maaloe2016auxiliary,sonderby2016ladder}. Similarly, GAN-based models ~\cite{qi2018global,salimans2016improved,zhao2018adversarial} also show promising results on many semi-supervised tasks, whose success is based on the adversarial learning. Below we review two ideas about semi-supervised learning that are closely related to the proposed model.
\par {\noindent\bf Consistent Predictions} One of the most popular and successful ideas in SSL is to encourage consistent predictions when inputs and/or models are perturbed. Inspired by Denoising Source Separation (DSS)~\cite{sarela2005denoising}, $\Gamma$-model~\cite{rasmus2015semi} applies a denoising layer to improve its performance by minimizing the impact of potential perturbations. $\Pi$-model~\cite{laine2016temporal} is further improved by adding stochastic augmentations on images and dropout~\cite{srivastava2014dropout} on network neurons to maximize prediction consistency. Furthermore, Virtual Adversarial Training (VAT)~\cite{miyato2018virtual} uses adversarial perturbations to replace the random noises in $\Pi$-model, making it more resilient against noises.

\par Compared with the previous methods on maximizing the prediction consistency under perturbations, the Mean Teacher model~\cite{tarvainen2017mean} updates the weights of a teacher model with an exponential moving average (EMA) of the weights from a sequence of student models as follows.
\begin{equation}
\label{eq:1}
\centering
   \Theta'_{\tau}=\alpha \Theta'_{\tau-1}+(1-\alpha)\Theta_{\tau}
\end{equation}
where 
$\Theta$ ($\Theta'$) is the weights of the Student (Teacher) Model, $\tau$ denotes the update step and $\alpha$ is a smoothing coefficient, which is always set to 0.999. This can result in stable and accurate predictions, and will be integrated into the proposed EnAET framework.

{\noindent\bf Confident Predictions} The other successful idea in SSL is to encourage a model to make confident predictions on both labeled and unlabeled data. For the feature space of a model, it is ideal that each class has a clear boundary with other classes. In other words, the boundary  should be far away from the high density regions of data. ``Pseudo-Label"~\cite{grandvalet2005semi}  implements this idea by minimizing the entropy loss of the predictions on unlabeled data. VAT~\cite{miyato2018virtual} also combines this entropy minimization term to make confident predictions. Similarly, several other works~\cite{lee2013pseudo,berthelot2019mixmatch} encourage confident predictions by constructing hard labels for high-confident unlabeled data to ``sharpen" the predictions.

\par Recently, MixUp~\cite{zhang2017mixup} was proposed to further improve the boundary between classes together with entropy minimization. Instead of only focusing on predicted results on given data, MixUp trains a model with the linear combination of the inputs and corresponding outputs. This has shown extraordinary performances in both supervised~\cite{zhang2017mixup} and semi-supervised tasks~\cite{berthelot2019mixmatch,verma2019interpolation}.

While current semi-supervised approaches is based on consistent prediction and confident prediction ideas, we find that self-supervised representations can successfully integrate with concurrent semi-supervised ideas by exploring the data variations under a variety of transformations. Unlike data augmentation applied to labeled data, we can self-train a semi-supervised model without relying on the labeled data. This can significantly boost the performances in semi-supervised as well as fully-supervised learning tasks. For this reason, we also review the related work on self-supervised methods below.
\subsection{Self-Supervised Learning}
Self-Supervised learning is also referred to as representation learning, which aims to learn representative features without label information. The learned features can then be frozen and used for classification. It can not only reduce the cost and time for collecting annotated data, but also successfully avoid the overfitting problem since it is trained without label information. Moreover, finetuning results~\cite{he2020momentum} with a pretrained model from self-supervised learning on different datasets also has shown its excellent encoding ability.
\par There are a wide variety of unsupervised models that apply different types of self-supervised signals to train deep networks, such as transformation parameters. Mehdi and Favaro~\cite{noroozi2016unsupervised} propose to solve Jigsaw puzzles to train a convolutional neural network.
Doersch et al.~\cite{doersch2015unsupervised} train the network by predicting the relative positions between sampled patches from an image as self-supervising information. Noroozi et al.~\cite{noroozi2017representation} count the features that satisfy equivalence relations between downsampled and tiled images, while Gidaris et al.~\cite{gidaris2018unsupervised} classify a discrete set of image rotations to train deep networks. Dosovitskiy et al.~\cite{dosovitskiy2014discriminative} create a set of surrogate classes from individual images. 
Unsupervised features have also been learned from videos by estimating the self-motion of moving objects between consecutive frames \cite{agrawal2015learning}.

More recently, Zhang et al.~\cite{zhang2019aet,qi2020learning} has demonstrated the state-of-the-art performances in many unsupervised tasks with the utilization of transformation information. It aims to learn a good representation of visual structures that can decode the transformations from the learned representations of original and transformed images. Inspired by this self-supervised idea, we develop a self-trained framework for semi-supervised tasks by exploring unlabeled data under a transformation ensemble.
\section{Ensemble AutoEncoding Transformations}
\label{sec:alleaet}
\begin{table*}[!htb]
\centering
\caption{Spatial Transformations.}
\label{tab:geot}
\begin{tabular}{ccccl}
\toprule
Name & DOF & Matrix & Effect & Property \\ \toprule
Projective & 8 & $\left[\begin{matrix}a_{1}  &  a_{2}  &  b_{1}\\a_{3}  &  a_{4}  &  b_{2} \\c_{1}  &  c_{2}  &  1\end{matrix}\right]$ & \begin{tabular}[c]{@{}c@{}}Translation+Rotation\\ +Scale+Aspect Ratio \\ + Shear+Projective\end{tabular} & \begin{tabular}[c]{@{}l@{}}Lines map to lines\\ Parallelism may not be maintained\\ Defined on the complement of line\end{tabular} \\\hline
Affine & 6 & $\left[\begin{matrix}a_{1}  &  a_{2}  &  b_{1} \\a_{3}  &  a_{4}  &  b_{2} \\0  &  0  &  1\end{matrix}\right]$ & \begin{tabular}[c]{@{}c@{}}Translation+Rotation \\ +Scale+Aspect Ratio \\ + Shear\end{tabular} & \begin{tabular}[c]{@{}l@{}}Preserves collinearity\&parallelism\\ Preserves the ratio of distances\\ Does not preserve angles or lengths\end{tabular} \\\hline
Similarity & 4 & $\left[\begin{matrix}a*cos(\theta)  &  -sin(\theta)  &  b_{1} \\sin(\theta) & a*cos(\theta)   &  b_{2} \\0  &  0 &  1\end{matrix}\right]$ & \begin{tabular}[c]{@{}c@{}}Translation\\+Rotation +Scale\end{tabular} & \begin{tabular}[c]{@{}l@{}}Preserves collinearity\&parallelism\\ Preserves general shape of objects\\ Preserves angles of objects\end{tabular} \\\hline
Euclidean & 3 & $\left[\begin{matrix}cos(\theta)  &  -sin(\theta)  &  b_{1} \\sin(\theta) & cos(\theta) &  b_{2} \\0  &  0  &  1\end{matrix}\right]$ & Translation+Rotation & \begin{tabular}[c]{@{}l@{}}Preserves collinearity\&parallelism\\ Preserves exact shape of objects\\ Preserves angles\&lengths of objects\end{tabular} \\ \bottomrule
\end{tabular}
\end{table*}
\par In this section, we introduce the proposed Ensemble AutoEncoding Transformation (EnAET) framework, a novel idea centered around leveraging an ensemble of spatial and non-spatial transformations to self-train  semi-supervised methods for improved performances. Although we choose MixMatch \cite{berthelot2019mixmatch} as the reference model, the EnAET framework is not limited to MixMatch.  Even though MixMatch has already achieved competitive performances over several datasets, we will demonstrate EnAET can still significantly further improve its performances as a general self-training mechanism.


\par
\begin{figure}[t]

\centering
%
\includegraphics[width=0.9\linewidth]{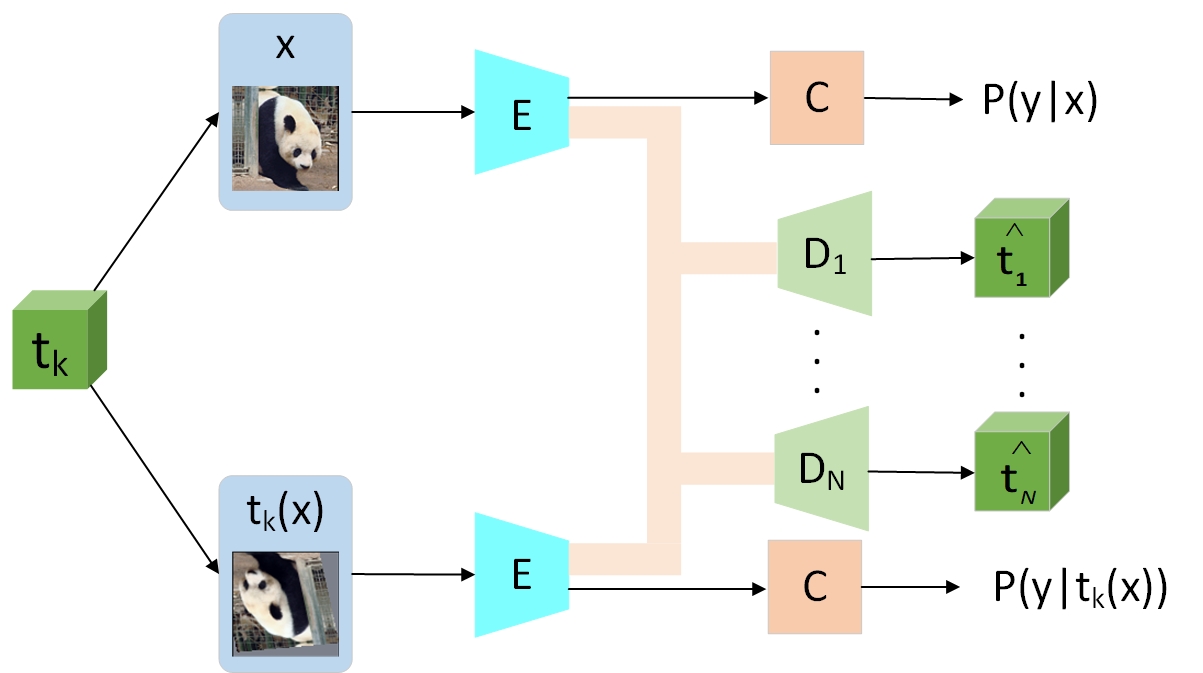}
\caption{The Ensemble Auto Encoding Transformation Pipeline.}
\label{fig:figAET}
\end{figure}

 Indeed, the difference between the features extracted from original and transformed images is caused by the applied transformations. Therefore, the transformation decoder can recover the transformations so long as the encoded features capture the necessary details of visual structures. Inspired by unsupervised work in \cite{zhang2019aet}, EnAET can self-train a good feature representation upon which a competitive semi-supervised classifier can be developed to explore an ensemble of spatial and non-spatial transformations. Since EnAET will not make use of labelled information, the EnAET framework can serve as a regularization part and integrate with all current semi-supervised methods and boost its performance. Fig.~\ref{fig:figAET} illustrates the EnAET framework. We will describe the details in Section \ref{sec:frameillustrate}.
\begin{algorithm*}[!htb]
\caption{Algorithms of Ensemble AutoEncoding Transformations (EnAET).}
\label{alg:A}
\begin{algorithmic}[1]
\Require a batch of labeled data pair $\mathcal{X}$, unlabeled data $\mathcal{U}$, the number of transformations in EnAET $N$, the balancing coefficients $\lambda_{\mathcal{U}'}$, $\lambda_{k}$, and $\gamma$.

\State $\mathcal{X'},\mathcal{U'}=MixMatch(\mathcal{X},\mathcal{U})$ \Comment{Use Algorithm 1 in  MixMatch~\cite{berthelot2019mixmatch}}
\State $\mathcal{L}_{SSL}=\mathcal{L}_{\mathcal{X}'}+\lambda_{\mathcal{U}'}*\mathcal{L}_{\mathcal{U}'}$ \Comment{Calculate SSL loss (see Eq.~(\ref{eq:3}))}
\For{$k$=1 to $N$}
\State $\mathcal{L}_{AET_{k}}=\mathbb E_{x\in \mathcal{U} ,t_k}~\|{D}\left[E(x),E(t_k(x))\right]-t_k\|^{2}$ \Comment{Calculate AET loss, which is illustrated in Section \ref{sec:aet}}.
\State $\mathcal L_{CL_k}=\mathbb E_{x\in \mathcal{U},t_k}\mathop\sum\limits_{y} P(y|x)\log\frac{P(y|x)}{P_{t_k}(y|x)}$ \Comment{Consistent Prediction loss, details in Section \ref{sec:consistencyaet}}
\EndFor
\State $\mathcal{L}=\mathcal{L}_{SSL}+\sum_{k=1}^{N}\lambda_{k}~\mathcal{L}_{AET_{k}}+\gamma~\sum_{k=1}^{N}\mathcal{L}_{CL_{k}}$ \Comment{Calculate the overall loss}
\State Apply $\mathcal{L}$ to update model.
\State Update teacher model: $\Theta'_{\tau}=\alpha \Theta'_{\tau-1}+(1-\alpha)\Theta_{\tau}$\Comment{Use EMA~\cite{tarvainen2017mean} to update the final model's loss (see Eq.~(\ref{eq:1}))}
\Ensure Student model with weight $\Theta$ and teacher model with weight $\Theta'$.
\end{algorithmic}
\end{algorithm*}
\subsection{Ensemble Transformations}
\label{sec:aet}

\par Zhang et al.~\cite{zhang2019aet} utilize the parameterized transformations as labels to train encoder to try to predict the most representative features. In the SSL setting, instead of pretraining the encoder as unsupervised learning, we formulate AutoEncoding Transformation (AET) loss as a regularization term along with the SSL loss to train classifiers.
\par In this study, we mainly focused on two types of transformations to self-train the SSL algorithms. We will briefly introduce it below.
{\noindent\bf Spatial Transformations} As introduced in \cite{hartley1999theory}, for any 2D spatial transformation, we can represent it with a matrix below in Eq.~(\ref{eq:geo}),
\begin{equation}
\centering
\label{eq:geo}
  \left[
 \begin{matrix}
   a_{1} & a_{2} & b_{1} \\
   a_{3} & a_{4} & b_{2} \\
   c_{1} & c_{2} & 1
  \end{matrix}
  \right] \left[
 \begin{matrix}
   x  \\
   y  \\
   1
  \end{matrix}
  \right]=\left[
 \begin{matrix}
   x'  \\
   y'  \\
   1
  \end{matrix}
  \right]
\end{equation}
where $\bigl( \begin{smallmatrix} a_{1} & a_{2}  \\
   a_{3} & a_{4} \\ \end{smallmatrix} \bigr)$ is a submatrix that controls the rotation, aspect ratio, shearing and scaling factors, $\bigl( \begin{smallmatrix} b_{1} \\
   b_{2} \\ \end{smallmatrix} \bigr)$ is the translation, and $\bigl( \begin{smallmatrix} c_{1}~
   c_{2} \\ \end{smallmatrix} \bigr)$ is the projection; $(x,y)$ is the coordinate of original image, while $(x',y')$ denotes the  coordinate after the transformation.
\begin{figure}[t]

\centering
%
\includegraphics[width=0.9\linewidth]{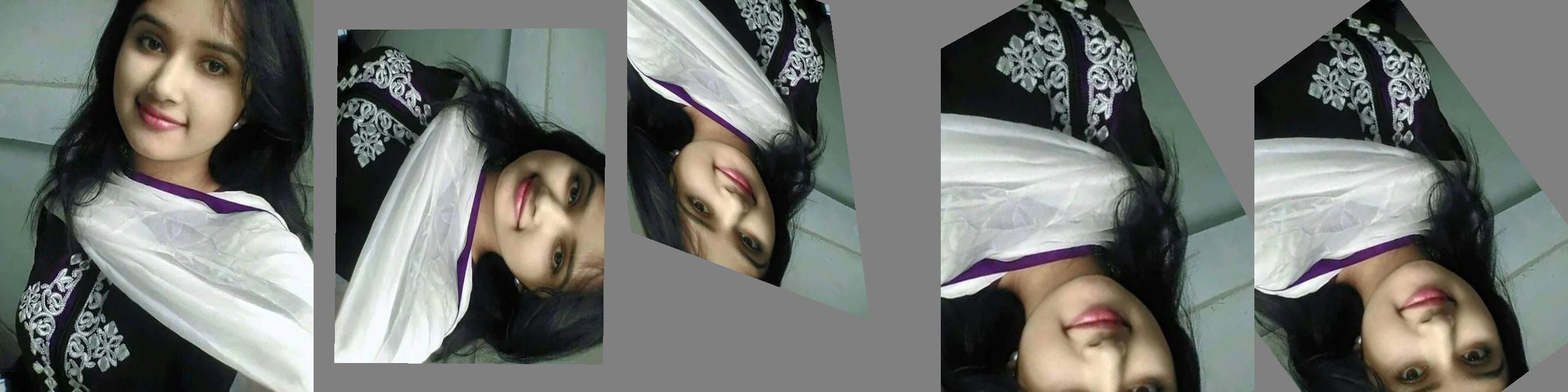}
\caption{Spatial transformations. The images are original, projective transformation, affine transformation, similarity transformation, euclidean transformation.}
\label{fig:figspatial}
\end{figure}
\par Based on Eq.~(\ref{eq:geo}), we incorporate four most representative transformations into the AET loss: 1) Projective transformation, 2) Affine transformation, 3) Similarity transformation, and 4) Euclidean transformation.  We illustrate and compare these transformations in Table~\ref{tab:geot} and Fig.~\ref{fig:figspatial}.  Although it seems that projective transformation includes the other transformations as its special cases, it is natural to raise the question of whether we still need to include the other transformations explicitly. However, noting the fact that the Euclidean, Similarity and Affine transformations rarely happen under Projective transformation when we use the random parameters to conduct projective transformation, we will explicitly sample these transformations in EnAET. Our ablation study also shows that the ensemble transformations can contribute to the model's performance.

\begin{figure}[t]

\centering
%
\includegraphics[width=0.9\linewidth]{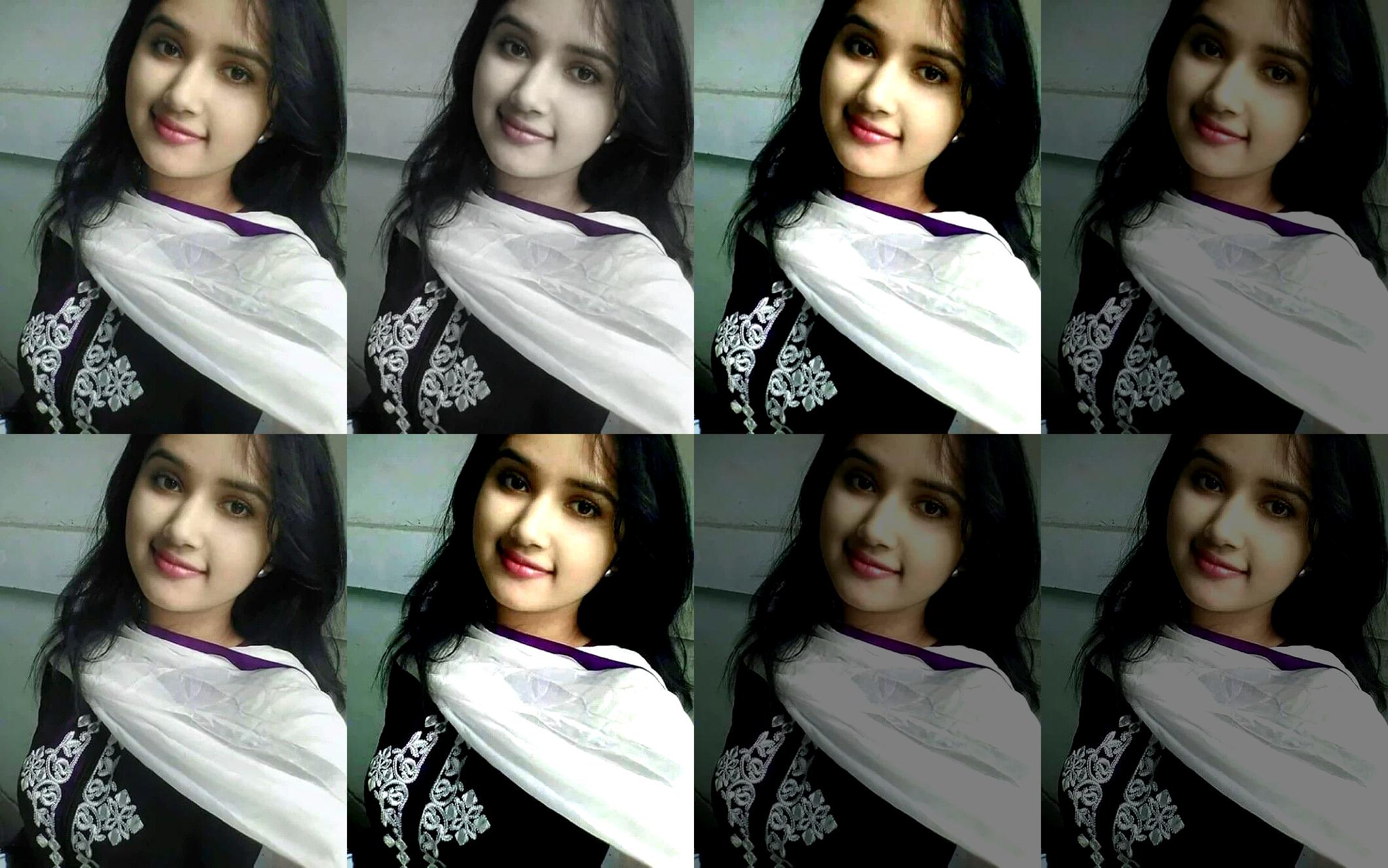}
\caption{Non-spatial transformations. The images are original, color transformation, contrast transformation, brightness transformation, sharpen transformation, color+contrast, color+contrast+brightness, color+contrast+brightness+sharpen.}
\label{fig:figCCBS}
\end{figure}

{\noindent\bf Non-spatial transformations} A good classifier can also recognize objects in different color, contrast, brightness, and sharpness conditions. Therefore, we also add these non-spatial transformations to EnAET framework. We consider four different non-spatial transformations as shown in Table~\ref{tab:CCBS}. For simplicity, these four transformations are applied as an entire non-spatial transformation with four strength parameters in EnAET. The effect of such a combined Color Contrast Brightness Sharpness (CCBS) transformation is illustrated in Fig.~\ref{fig:figCCBS}.

\begin{table}[]
\centering
\caption{Non-spatial Transformations.}
\label{tab:CCBS}
\begin{tabular}{ll}
\toprule
Transform & Description \\ \midrule
Color & Adjusts color balance of image \\\hline
Contrast & \begin{tabular}[c]{@{}l@{}}Adjusts difference of light pixels and dark pixels in image\end{tabular} \\\hline
Brightness & \begin{tabular}[c]{@{}l@{}}Adds or subtracts to= image matrix to change brightness\end{tabular} \\\hline
Sharpness & \begin{tabular}[c]{@{}l@{}}Adjusts pixels to make image appear sharper \end{tabular} \\ \bottomrule
\end{tabular}
\end{table}

To apply AutoEncoding Transformation (AET) idea to self-train the encoder part, we can formulate AET loss in Eq.~(\ref{eq:aetloss}).
\begin{equation}
\centering
\label{eq:aetloss}
  \mathcal L_{AET_k}=\mathbb E_{x,t_k}~\|{D}\left[E(x),E(t_k(x))\right]-t_k\|^{2}
\end{equation}
where ${D}$ denotes the transformation decoder, $E$ represents the encoder, and $t_k$  is the sampled transformation of type $k$. The AET loss computes the Mean-Squared Error (MSE) between the predicted transformation and the sampled transformation.
\par Then, we can minimize a linear combination of the SSL loss and the AET loss to train a classifier over the network weights $\Theta$
\begin{equation}
\centering
\min_{\Theta}~\mathcal L_{SSL} + \sum_{k=1}^N \lambda_k\mathcal L_{AET_k}
\end{equation}
where $\lambda_k$ is the weight on the AET loss $\mathcal L_{AET_k}$ for the transformation $t_k$ of the $k$th type.

Here, the SSL loss $\mathcal L_{SSL}$ can be any loss used to self-train a semi-supervised classifier in literature. Particularly, we will use the MixMatch loss \cite{berthelot2019mixmatch} that yields the state-of-the-art SSL result. In other words, we use MixMatch as the baseline to demonstrate the proposed framework can further improve an even very competitive model.


\subsection{Making Consistent Predictions}
\label{sec:consistencyaet}


\par Inspired by the common SSL algorithms, transformation invariance by making consistent predictions on image labels under transformations \cite{rifai2011contractive,miyato2018virtual} are widely used.
\par Similarly, apart from the AET loss for our ensemble transformations, we also explore the consistency predictions for our transformations. To achieve this, we propose our consistency loss (CL) by minimizing the KL divergence  between the ``guessed label" $P(y|x)$ on an original image $x$ and $P_{t}(y|x)\triangleq P(y|t(x))$ on a transformed image $t(x)$
\begin{equation}
\centering
\label{eq:kl}
  \mathcal L_{CL}=\mathbb E_{x,t}\mathop\sum\limits_{y} P(y|x)\log\frac{P(y|x)}{P_{t}(y|x)}
\end{equation}
to make consistent predictions under different transformations, where the expectation is taken over the sampled data and transformations.


Furthermore, since we used MixMatch as our baseline of the framework, we use sharpened ``guessed label" $P(y|x)$ for original image $x$.
\begin{table*}[!htp]
\centering
\caption{Error rates of different models on CIFAR-10.}
\label{tab:cifar10}
\begin{tabular}{lccccccc}
\toprule
Methods/Labels & 50 & 100 & 250 & 500 & 1000 & 2000 & 4000 \\ \midrule
{$\Pi$-}Model \cite{laine2016temporal,sajjadi2016regularization} & -- & -- & 53.02$\pm$2.05 & 41.82$\pm$1.52 & 31.53$\pm$0.98 & 23.07$\pm$0.66 & 17.41$\pm$0.37 \\
PseudoLabel \cite{lee2013pseudo} & -- & -- & 49.98$\pm$1.17 & 40.55$\pm$1.70 & 30.91$\pm$1.73 & 21.96$\pm$0.42 & 16.21$\pm$0.11 \\
MixUp \cite{zhang2017mixup} & -- & -- & 47.43$\pm$0.92 & 36.17$\pm$2.82 & 25.72$\pm$0.66 & 18.14$\pm$1.06 & 13.15$\pm$0.20 \\
VAT \cite{miyato2018virtual} & -- & -- & 36.03$\pm$2.82 & 26.11$\pm$1.52 & 18.68$\pm$0.40 & 14.40$\pm$0.15 & 11.05$\pm$0.31 \\
 MeanTeacher \cite{tarvainen2017mean} & -- & -- & 47.32$\pm$4.71 & 42.01$\pm$5.86 & 17.32$\pm$4.00 & 12.17$\pm$0.22 & 10.36$\pm$0.25 \\
MixMatch \cite{berthelot2019mixmatch} & -- & -- & 11.08$\pm$0.87 & 9.65$\pm$0.94 & 7.75$\pm$0.32 & 7.03$\pm$0.15 & 6.24$\pm$0.06 \\
\midrule
EnAET & \textbf{16.45} & \textbf{9.35} & \textbf{7.6$\pm$0.34} & \textbf{7.27} & \textbf{6.95} & \textbf{6.00} & \textbf{5.35} \\ \bottomrule
\end{tabular}
\end{table*}

\begin{table}[]
\centering
\caption{Error rates of different models on CIFAR-100.}
\label{tab:cifar100}
\begin{tabular}{lccc}
\toprule
Methods/Labels & 1000 & 5000 & 10000 \\ \midrule
Supervised-only & -- & -- & 51.21$\pm$0.33 \\
$\Pi$-Model \cite{sajjadi2016regularization} & -- & -- & 39.19$\pm$0.36 \\
 Temporal ensembling \cite{laine2016temporal} & -- & -- & 38.65$\pm$0.51 \\\midrule
EnAET & \textbf{58.73} & \textbf{31.83} & \textbf{26.93$\pm$0.21} \\ \bottomrule
\end{tabular}
\end{table}

\subsection{Overall Framework}
\label{sec:frameillustrate}
\par We illustrate the framework of the proposed EnAET in Fig.~\ref{fig:figAET}. For each image $x$, we apply five different transformations: $t_{1}$(Projective),
$t_{2}$(Affine),
$t_{3}$(Similarity),
$t_{4}$(Euclidean),
$t_{5}$(CCBS).
\par After that, the network is split into three parts: an representation encoder $E$, a classifier $C$, and a set of decoders $D_k$ each for a type of transformation $t_k$. The original input  $x$ and all its transformed counterparts $t_{k}(x)$ are fed through the network.  The original and transformed images have a Siamese encoder $E$ and classifier $C$ with shared weights.

\par As in the baseline MixMatch settings, we use Wide ResNet-28-2 network as our architecture. It consists of four blocks, and we use the last block as the classifier $C$, while the other three blocks constitute the encoder $E$. Also, all the decoders $D_{k}$'s share the same network architecture as the classifier $C$ but with different weights.

\par The representations of the original and transformed images will be concatenated to predict the  parameters of each transformation $t_k$ by the corresponding decoder $D_{k}$.

\par The classifier $C$ is built upon the encoded representation to output the label predictions $P({y|x})$ and $P({y|t(x)})$ for both the original and transformed images, respectively. Following our baseline MixMatch, we used the sharpened ``guessed label" as $P({y|x})$. The label prediction of original image needs to be ``sharpened" \cite{goodfellow2016deep} to reach a high degree of prediction confidence by minimizing the prediction entropy.
\par After splitting a network into Encoder $E$ and Classifier $C$, we can then update the network with a linear combination of three losses: semi-supervised loss $\mathcal{L}_{SSL_{k}}$, AET loss $\mathcal{L}_{AET_{k}}$ and consistency loss $\mathcal{L}_{CL_{k}}$. Our framework can be illustrated in Algorithm ~\ref{alg:A}. It is  clear that our framework is independent of any semi-supervised learning algorithm. That is to say, for lines 1-2 in Algorithm~\ref{alg:A}, we can easily replace them with any SSL algorithms to apply the framework when new SSL methods are proposed.
\begin{table}[]
\centering
\caption{Error rates of different models on STL10.}
\label{tab:stl10}
\begin{tabular}{lcc}
\toprule
Methods/Labels & 1000 & 5000 \\\midrule
CutOut \cite{devries2017improved} & - & 12.74 \\
IIC \cite{ji2018invariant} & - & 11.20 \\
SWWAE \cite{zhao2015stacked} & 25.70 & - \\
CC-GAN$^2$ \cite{denton2016semi} & 22.20 & - \\
MixMatch \cite{berthelot2019mixmatch} & 10.18$\pm$1.46 & 5.59 \\\midrule
EnAET & \textbf{8.04} & \textbf{4.52}\\
\bottomrule
\end{tabular}
\end{table}

\subsection{SSL Baseline: MixMatch}
\par MixMatch~\cite{berthelot2019mixmatch} is the current state-of-the-art for the semi-supervised classification task. It uses many ideas in previous approaches to build a new algorithm. Among them are  Mixup~\cite{zhang2017mixup} by training a network with a convex combination of examples and their labels, as well as entropy minimization by sharpening the label predictions. In addition, a guessed label with data augmentation also contributes to more consistent label predictions.

Formally, Algorithm~1 in MixMatch~\cite{berthelot2019mixmatch} mixes up a batch of labeled $\mathcal X$ and unlabeled $\mathcal U$ to obtain the mixed-up $\mathcal X^\prime,\mathcal U^\prime$. Then, it minimizes the following SSL loss $\mathcal L_{SSL}$ to train its model,
\begin{equation}
\label{eq:3}
\centering
\begin{cases}
   \mathcal{L}_{\mathcal{X}'}=\mathbb E_{(x,y)\in \mathcal{X'}} {H}(y, f(x,\Theta))\\
   \mathcal{L}_{\mathcal{U}'}=\mathbb E_{(u,q)\in \mathcal{U'}}||f(u,\Theta)-q||^{2}
   \\
   \mathcal{L}_{SSL}=\mathcal{L}_{\mathcal{X}'}+\lambda_{\mathcal{U}'}\mathcal{L}_{\mathcal{U}'}
\end{cases}
\end{equation}
where $\mathcal{X}'$ and $\mathcal{U}'$ are the labeled and unlabeled data resulting from MixMatch, $H$ is the cross-entropy between two distributions, and $q$ is the sharpened predictions on a unlabeled sample $u$ for each $(u,q)\in\mathcal U'$.
\begin{table*}[!htb]
\centering
\caption{Error rates of different models on SVHN.}
\label{tab:svhn}
\begin{tabular}{lcccccc}
\toprule
Methods/Labels & 100 & 250 & 500 & 1000 & 2000 & 4000 \\ \midrule
$\Pi$-Model \cite{laine2016temporal,sajjadi2016regularization}& -- & 17.65$\pm$0.27 & 11.44$\pm$0.39 & 8.6$\pm$0.18 & 6.94$\pm$0.27 & 5.57$\pm$0.14 \\
 PseudoLabel \cite{lee2013pseudo}& -- & 21.16$\pm$0.88 & 14.35$\pm$0.37 & 10.19$\pm$0.41 & 7.54$\pm$0.27 & 5.71$\pm$0.07 \\
MixUp \cite{zhang2017mixup}& -- & 39.97$\pm$1.89 & 29.62$\pm$1.54 & 16.79$\pm$0.63 & 10.47$\pm$0.48 & 7.96$\pm$0.14 \\
 VAT \cite{miyato2018virtual}& -- & 8.41$\pm$1.01 & 7.44$\pm$0.79 & 5.98$\pm$0.21 & 4.85$\pm$0.23 & 4.20$\pm$0.15 \\
 MeanTeacher \cite{tarvainen2017mean}& -- & 6.45$\pm$2.43 & 3.82$\pm$0.17 & 3.75$\pm$0.10 & 3.51$\pm$0.09 & 3.39$\pm$0.11 \\
 MixMatch \cite{berthelot2019mixmatch}& -- & 3.78$\pm$0.26 & 3.64$\pm$0.46 & 3.27$\pm$0.31 & 3.04$\pm$0.13 & 2.89$\pm$0.06 \\\midrule
EnAET & \textbf{16.92} & \textbf{3.21$\pm$0.21} & \textbf{3.05} & \textbf{2.92} & \textbf{2.84} & \textbf{2.69} \\ \bottomrule
\end{tabular}
\end{table*}

\begin{table}[]
\centering
\caption{Comparison of error rates with Wide ResNet-28-2-Large.}
\label{tab:largesemi}
\begin{tabular}{lccc}
\toprule
Methods/Labels & \begin{tabular}[c]{@{}c@{}}CIFAR-10\\ 4000 label\end{tabular} & \begin{tabular}[c]{@{}c@{}}CIFAR-100\\ 10000 label\end{tabular} & \begin{tabular}[c]{@{}c@{}}SVHN\\ 1000 label\end{tabular} \\ \midrule
Mean Teacher \cite{tarvainen2017mean} & 6.28 & -- & -- \\
SWA \cite{athiwaratkun2018improving} & 5.00 & 28.80 & -- \\
Fast SWA \cite{athiwaratkun2018improving} & 5.0 & 28.0 & -- \\
MixMatch \cite{berthelot2019mixmatch} & 4.95$\pm$0.08 & 25.88$\pm$0.30 & -- \\\midrule
EnAET & \textbf{4.18} & \textbf{22.92} & \textbf{2.42} \\ \bottomrule
\end{tabular}
\end{table}

\section{Experiments}
\label{sec:experiments}
\begin{table}[]
\centering
\caption{Error Rates of fully supervised models with a Wide ResNet-28-2 backbone.}
\label{tab:wideresnet2}
\begin{tabular}{lccc}
\toprule
Methods/Labels & CIFAR-10 & CIFAR-100 & SVHN \\ \midrule
Baseline \cite{zagoruyko2016wide} & 5.1 & 24.6 & 3.3 \\
AutoAugment \cite{cubuk2018autoaugment} & 4.1 & 21.5 & 1.7 \\
MixMatch \cite{berthelot2019mixmatch} & 4.13 & -- & 2.59 \\\midrule
EnAET & \textbf{3.55} & \textbf{20.55} & \textbf{2.48} \\ \bottomrule
\end{tabular}
\end{table}

\subsection{Training Details}
For a fair comparison with other SSL methods, we follow our baseline MixMatch setttings and use ``Wide ResNet-28-2" as our backbone network and a ``Wide ResNet-28-2" architecture  with 135 filters per layer as our larger model, which we will refer to as ``Wide ResNet-28-2-Large" in the following. This also follows the evaluation setup for the baseline methods in \cite{oliver2018realistic}.

\par For our encoder $E$ and $C$, We follow our baseline MixMatch's setting and use the Adam solver \cite{kingma2014semi} with a learning rate of 0.002 to train. For decoder $D_{k}$, the SGD optimizer is used  with an initial learning rate of 0.1. Then we use a cosine \cite{loshchilov2016sgdr} scheduler for a learning rate decay from 0.1 to 0.0001. We also fix the weight decay rate to 5e-4.

\par As settings in \cite{berthelot2019mixmatch}, for all experiments, we use a batch size  of $128$ images. The model is trained for $1024$ epochs. For the sake of fair comparison,  the mean error rate of the last 20 models is reported . Also, we report the error variance based on 4 runs with different random seeds.

\par We also adopt the same hyper-parameter setup to minimize the MixMatch loss as in \cite{berthelot2019mixmatch}.
The weight $\lambda_{k}$ of the AET loss is initialized to $1.0, 0.75, 0.5, 0.2, 0.05$ for the four spatial transformations and the CCBS transformation, respectively, and we fix that for different datasets.  The weight  $\gamma$ of the CL loss is also always set to $0.2$
as an initial value, and we found it would not influence the overall performance too much, which only needs to be slightly adjusted if you want to transfer to more datasets.
Like our baseline MixMatch~\cite{berthelot2019mixmatch}, we use a warm-up strategy for the weights of these losses.

\par Finally, we also use the exponential moving average (EMA)~\cite{tarvainen2017mean} in the experiments with a rate of $0.999$.

\subsection{Transformation Details}
\par For spatial transformations, considering they can all be expressed by a matrix shown in Eq.~(\ref{eq:geo}), we use the same operation settings for all of them. For random rotation, we set the rotation degree from [\ang{-180},\ang{180}]. For the translation factor, we randomly sample the translation from [-0.2,0.2] for both horizontal and vertical directions. Also, we sample the scaling factor from [0.8,1.2] to make the scaled image fall in a proper range. With the shearing factor, we limit the shearing in [\ang{-30},\ang{30}] to make sure the image can still be recognized. For the projective factor, we set the translation factor for the 4 corners of an image in [-0.125,0.125] for both horizontal and vertical directions.
\par For non-spatial transformations, we randomly sample the magnitude for color, contrast, brightness and sharpness from [0.2,1.8] to keep the transformed images recognizable by human beings.
\begin{table}[]
\vspace{-0.5cm}
\centering
\caption{Error Rates of fully supervised models with Wide ResNet-28-2-Large.}
\label{tab:largefull}
\begin{tabular}{lccc}
\toprule
Methods/Labels & CIFAR-10 & CIFAR-100 & SVHN \\ \midrule
Baseline \cite{zagoruyko2016wide} & 3.9 & 18.8 & 3.1 \\
AutoAugment \cite{cubuk2018autoaugment} & 2.6 & 17.1 & \textbf{1.9} \\
PBA \cite{ho2019population} & 2.6 & 16.7 & - \\
Fast AA \cite{lim2019fast} & 2.7 & 17.3 & - \\
EnAET & \textbf{1.99} & 16.87 & 2.22 \\ \midrule
ProxylessNAS \cite{cai2018proxylessnas} & 2.08 & - & - \\
CutMix \cite{yun2019cutmix} & 2.88 & \textbf{13.81} & - \\ \bottomrule
\end{tabular}
\vspace{-0.5cm}
\end{table}
\subsection{EnAET for Semi-Supervised Learning}
\label{sec:ssl}
To evaluate the proposed method, we perform semi-supervised tasks on four datasets: CIFAR-10 and CIFAR-100 \cite{krizhevsky2009learning}, SVHN \cite{netzer2011reading}, and STL-10 \cite{coates2011analysis}.
\subsubsection{CIFAR-10 Results}
 For CIFAR-10, we evaluate the compared methods with different sizes of labeled data.  The results are reported in Table~\ref{tab:cifar10}. Experiments are conducted 4 times with different random seeds to test the stability of our method and we find out that our method's variance is very small when we only use 250 labels. Therefore, we report the variance based on 4 runs for 250 labels and do not further perform more experiments for other conditions.

\par The results show that the proposed framework outperforms all the state-of-the-art methods. For example, the proposed model achieves a 7.6\% error rate with $250$ labels compared with  previous best rate of 11.08\%, which is our baseline SSL algorithm. Here we reduced around relative 2.5\% error rate compared to SSL baseline.  Notably, we are the first to conduct experiments with only $50$ labels and $100$ labels, and achieve 16.45\% and 9.35\% error rates, respectively. It is worth noting that the proposed framework with 50 labels even outperforms most methods with $1,000$ labels.

\subsubsection{CIFAR-100 Results}

 Different models are compared in Table~\ref{tab:cifar100} on CIFAR-100.  The proposed EnAET achieves an error rate of 58.73\% and 31.83\% with only $1,000$ and $5,000$ labels. The performance of EnAET with $5,000$ labels is even better than other models with $10,000$ labels. Here we do not have our baseline experiments' information, but our advantage compared to previous SSL algorithms is very obvious.

\subsubsection{SVHN Results}
 Compared with the previous methods, the proposed model achieves a new  performance record on the SVHN dataset as shown in Table~\ref{tab:svhn}. Notably, we are the first to test SVHN under 100 images and achieved 16.92\% error rate, which is even better compared to some methods with $1,000$ labels. Compared to our baseline with very low error rate, we can still have around 0.5\% accuracy improvement.

\subsubsection{STL10 Results}
STL-10 contains $5,000$ labeled images and $100,000$ unlabeled images and is specifically designed for semi-supervised task.
Follow the same setting as baseline MixMatch, we achieve the best performance with $1,000$ and $5,000$ labeled images in Table~\ref{tab:stl10}. We reduce the error rate from 10.18\% to 8.04\% with $1,000$ labeled images, and set a record error rate of 4.52\% when using all labeled data.

\subsubsection{Performance of Wide ResNet-28-2-Large}
Following the settings in MixMatch, we also evaluate our framework with larger architecture. The results on 3 different classical settings, we have clearly improved compared to our baseline MixMatch. For instance, we reduced error rate from 25.88\% to 22.92\% with around 3\% improvement under CIFAR-100 with 10000 labels settings.

\subsection{Supervised Learning}
\label{sec:sl}
We also compare different models in the fully supervised setting on all dataset by using all labeled data. We show that the proposed framewrok can still achieve the best results with the same network architecture.
\subsubsection{Wide ResNet-28-2 Backbone}

We compare different methods by using Wide ResNet-28-2 as the backbone in Table~\ref{tab:wideresnet2}. It's clear our framwork can achieve best performance and can still have some improvement compared to MixMatch though we used all labels now.

\subsubsection{Wide ResNet-28-2-Large Backbone} We also compare the proposed framework with the current state-of-the-art methods to demonstrate its remarkable performance in Table~\ref{tab:largefull} though the architecture is less complex compared to other methods. Compared to baseline, we have improvment with a large margin. It is worth noting EnAET achieves the state of the art on CIFAR-10 with a less complicated architecture.

\subsection{Ablation Study}
\label{sec:abs}
\begin{table}[!htb]
\centering
\caption{Ablation study of EnAET on CIFAR-10.}
\label{tab:ablation}
\begin{tabular}{cc}
\toprule
Ablation & 250labels \\ \midrule
EnAET & \textbf{7.64}\\
Only Projective Transformation & 9.96 \\
Only Affine Transformation & 8.12 \\
Only Similarity Transformation & 10.25 \\
Only Euclidean Transformation & 10.56 \\
Only CCBS Transformation & 15.34\\
Remove CL loss & 8.39 \\
Remove AET loss & 13.54 \\
Baseline: MixMatch & 11.08\\\bottomrule
\end{tabular}
\end{table}
We perform an ablation study of the proposed EnAET on CIFAR-10 and results are shown in Table~\ref{tab:ablation}. Specifically, we focus on the roles of different transformations  and the different components in EnAET:
\par 1) The ensemble of transformations can greatly contribute to the performance compared with any single transformation.
\par 2) The consistency loss can not contribute much to the performance but can increase the training stability.
\par 3) The AET loss regularization is the key to boosting the performance on top of the SSL baseline.

\section{Conclusion}
\label{sec:conclusion}
In this paper, we present EnAET, a novel  framework that integrates the idea of self-trained representations into semi-supervised learning models. Throughout our experiments, the proposed method outperforms the state-of-the-art methods on all the datasets by a significant margin  and greatly boost the performance of the semi-supervised baselines. Moreover, with the same architecture, the proposed method can also greatly improve the supervised learning baselines. Furthermore, on many datasets, EnAET even performs better than ProxylessNAS, the auto-augmentation policy search with a less complicated architecture. In the future, we will employ policy search to find the best combination of transformations and their ranges, which we believe can lead to even more competitive results.


%


\ifCLASSOPTIONcaptionsoff
  \newpage
\fi

\bibliographystyle{IEEEtran}
\bibliography{egbib}

%



%

\begin{IEEEbiography}[{\includegraphics[width=1in,height=1.25in,clip,keepaspectratio]{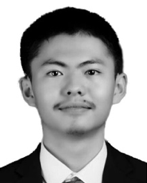}}]{Xiao Wang}
Xiao Wang received his B.S. degree from Department of Computer Science in Xi’an Jiaotong University, Xi’an, China in 2018. He is currently pursuing his Ph.D. degree in the Department of Computer Science at Purdue University, West Lafayette, IN, USA. His research interests include deep learning, computer vision, bioinformatics and intelligent systems.
\end{IEEEbiography}

\begin{IEEEbiography}[{\includegraphics[width=1in,height=1.25in,clip,keepaspectratio]{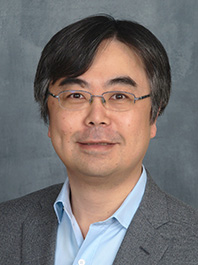}}]{Daisuke Kihara}
Daisuke Kihara is professor of Department of Biological Sciences and Department of Computer Science at Purdue University, West Lafayette, Indiana, USA. He has received B.S.
degree from University of Tokyo, Japan in 1994, M.S. and
Ph.D. degree from Kyoto University, Japan in 1996 and 1999, respectively. His research area is protein bioinformatics, which include
protein structure, docking, and function prediction. He
is named Showalter University Faculty Scholar from
Purdue University in 2013.
\end{IEEEbiography}

\begin{IEEEbiography}[{\includegraphics[width=1in,height=1.25in,clip,keepaspectratio]{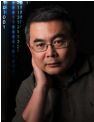}}]{Jiebo Luo}
Jiebo Luo (S93, M96, SM99, F09) is a Professor of Computer Science at the University of Rochester which he joined in 2011 after a prolific career of fifteen years at Kodak Research Laboratories. He has authored nearly 500 technical papers and holds over 90 U.S. patents. His research interests include computer vision, NLP, machine learning, data mining, computational social science, and digital health. He has been involved in numerous technical conferences, including serving as a program co-chair of ACM Multimedia 2010, IEEE CVPR 2012, ACM ICMR 2016, and IEEE ICIP 2017, as well as a general co-chair of ACM Multimedia 2018. He has served on the editorial boards of the IEEE Transactions on Pattern Analysis and Machine Intelligence (TPAMI), IEEE Transactions on Multimedia (TMM), IEEE Transactions on Circuits and Systems for Video Technology (TCSVT), IEEE Transactions on Big Data (TBD), ACM Transactions on Intelligent Systems and Technology (TIST), Pattern Recognition, Knowledge and Information Systems (KAIS), Machine Vision and Applications, and Journal of Electronic Imaging. He is the current Editor-in-Chief of the IEEE Transactions on Multimedia. Professor Luo is also a Fellow of ACM, AAAI, SPIE, and IAPR.
\end{IEEEbiography}

\begin{IEEEbiography}[{\includegraphics[width=1in,height=1.25in,clip,keepaspectratio]{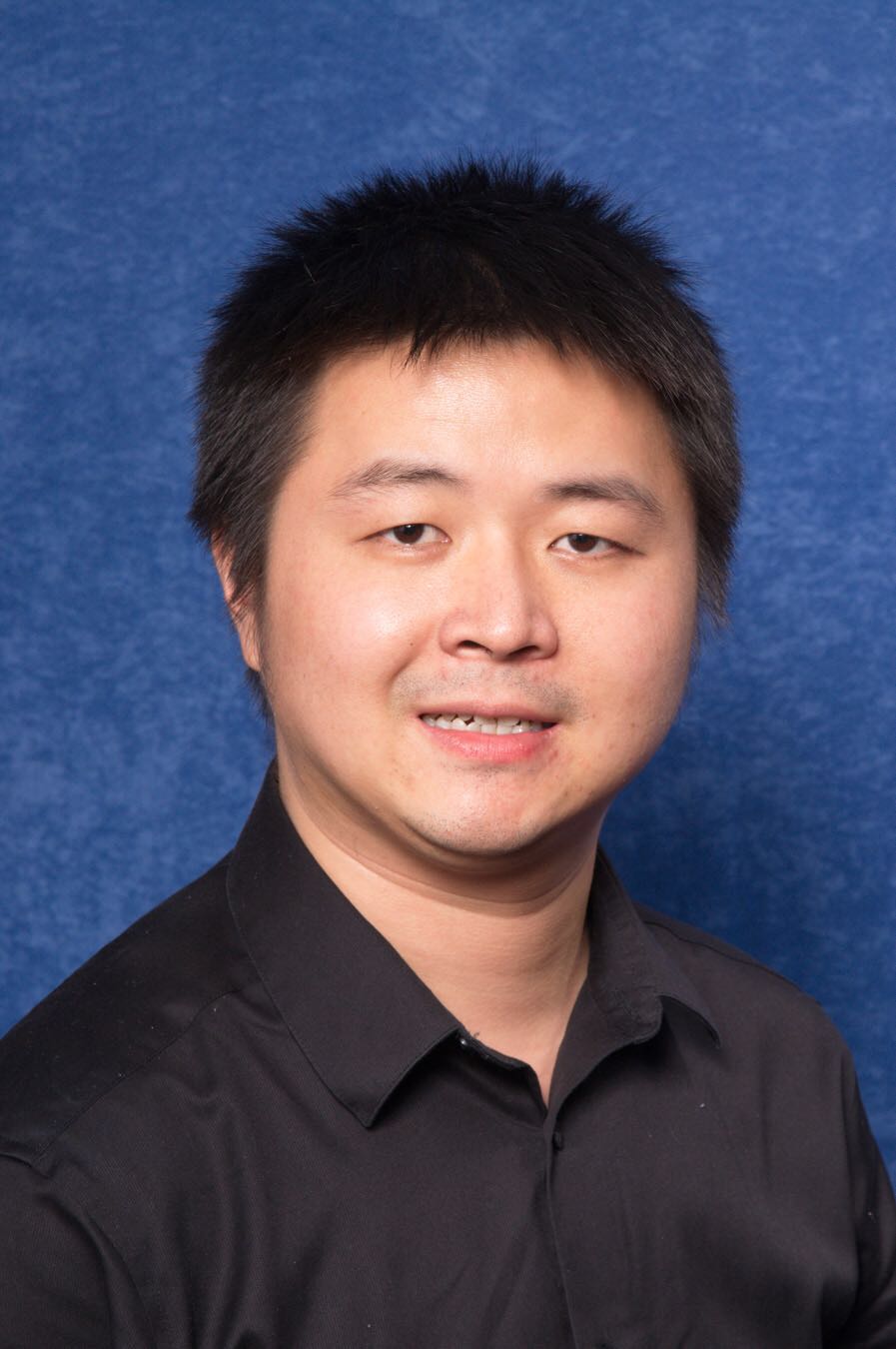}}]{Guo-Jun Qi}
Guo-Jun Qi (M14-SM18) is the
Chief Scientist leading and overseeing an international R\&D team for multiple artificial intelligent
services on the Huawei Cloud since August
2018. He was a faculty member in the Department
of Computer Science and the director of
MAchine Perception and LEarning (MAPLE) Lab
at the University of Central Florida since August
2014. Prior to that, he was also a Research Staff
Member at IBM T.J. Watson Research Center,
Yorktown Heights, NY. His research interests include
machine learning and knowledge discovery from multi-modal data
sources to build smart and reliable information and decision-making
systems. Dr. Qi has published more than 100 papers in a broad range
of venues in pattern recognition, machine learning and computer vision.
He also has served or will serve as a general co-chair for ICME 2021,
technical program co-chair for ACM Multimedia 2020, ICIMCS 2018
and MMM 2016, as well as an area chair (senior program committee
member) for multiple academic conferences. Dr. Qi is an associate editor
for IEEE Transactions on Circuits and Systems for Video Technology (TCSVT),
IEEE Transactions on Multimedia (T-MM), IEEE Transactions on
Image Processing (T-IP), Pattern Recognition (PR), and ACM Transactions
on Knowledge Discovery from Data (T-KDD).
\end{IEEEbiography}








\end{document}